\documentclass{article} 
 \usepackage[preprint]{neurips_arxiv_2020}

\usepackage{amsmath,amsfonts,bm}

\def\eqref#1{equation~\ref{#1}}

\def\1{\bm{1}}

\DeclareMathAlphabet{\mathsfit}{\encodingdefault}{\sfdefault}{m}{sl}
\SetMathAlphabet{\mathsfit}{bold}{\encodingdefault}{\sfdefault}{bx}{n}

\usepackage{hyperref}
\usepackage{url}
\usepackage{multirow}
\usepackage{booktabs}
\usepackage{graphicx}
\usepackage{chngpage,comment}
\usepackage[disable]{todonotes}
\usepackage{amsmath}

\title{Profile Prediction: An Alignment-Based \\ Pre-Training Task for Protein Sequence Models}

\author{Pascal Sturmfels\thanks{This research was conducted during the author’s internship at Salesforce Research} \\
Paul G. Allen School of Computer Science\\
University of Washington\\
\texttt{psturm@cs.washington.edu} \\
\And
Jesse Vig  \\
Salesforce Research \\
\texttt{jvig@salesforce.com} \\
\And
Ali Madani  \\
Salesforce Research \\
\texttt{amadani@salesforce.com} \\
\And
Nazneen Fatema Rajani \\
Salesforce Research \\
\texttt{nazneen.rajani@salesforce.com} \\
}

\begin{document}

\maketitle

\begin{abstract}
For protein sequence datasets, unlabeled data has greatly outpaced labeled data due to the high cost of wet-lab characterization. Recent deep-learning approaches to protein prediction have shown that pre-training on unlabeled data can yield useful representations for downstream tasks. However, the optimal pre-training strategy remains an open question. Instead of strictly borrowing from natural language processing (NLP) in the form of masked or autoregressive language modeling, we introduce a new pre-training task: directly predicting protein profiles derived from multiple sequence alignments. Using a set of five, standardized downstream tasks for protein models, we demonstrate that our pre-training task along with a multi-task objective outperforms masked language modeling alone on all five tasks. Our results suggest that protein sequence models may benefit from leveraging biologically-inspired inductive biases that go beyond existing language modeling techniques in NLP.
\end{abstract}

\section{Introduction}

Experimental measurements of protein properties, ranging from structural to functional characterization, are usually far more expensive than gathering unlabeled protein sequence data \citep{rocklin2017global, stevens2003cost}.
Predicting protein properties directly from sequence data using machine learning models could potentially enable a new era of biological discovery.
However, the vast majority of existing protein sequence data is unlabeled, lacking both structural (e.g. x-ray crystallography) or functional (e.g. enzymatic activity) labels.
In other domains, such as computer vision or natural language processing (NLP), existing approaches leverage large, unlabeled datasets via self-supervised pre-training: the practice of first training a model using a loss function derived solely from unlabeled data.
In the protein domain, self-supervision has recently been successful at improving downstream performance on canonical tasks in protein science and engineering \citep{devlin2019bert, vaswani2017attention, rao2019evaluating}. 

There  are  several  similarities between protein sequence modeling and NLP---namely, sequences comprised of a discrete set of characters as input, and far more unlabeled data than labeled. This has inspired many recent papers adapting NLP models to protein sequence tasks \citep{alley2019unified, elnaggar2020prottrans, uniprot2019uniprot, madani2020progen, rives2019biological, strodthoff2020udsmprot, nambiar2020transforming}. These papers employ not only architectures inspired by NLP but also pre-training objectives: masked language modeling and auto regressive generation. Unfortunately, on some tasks, such as secondary structure and contact prediction, purely using masked language modeling objectives for pretraining has only marginally improved performance as compared to standard alignment-based techniques \citep{rao2019evaluating}. This presents an opportunity to design new pre-training tasks and inductive biases that are better tailored toward the underlying protein biology.

Some existing work has already begun to investigate protein-specific pre-training tasks. \cite{bepler2018learning} propose a new task called soft symmetric alignment (SSA), which defines a custom similarity function that compares the pairwise distances between protein embeddings. \cite{Lu2020.09.04.283929} introduce a contrastive loss that trains an RNN to discriminate fragments coming from a source sequence versus randomly sampled fragments from other sequences. Our work differs from both existing works in that we make direct use of multiple sequence alignments (MSAs).

\begin{figure*}
    \centering
    \includegraphics[width=0.55\paperwidth]{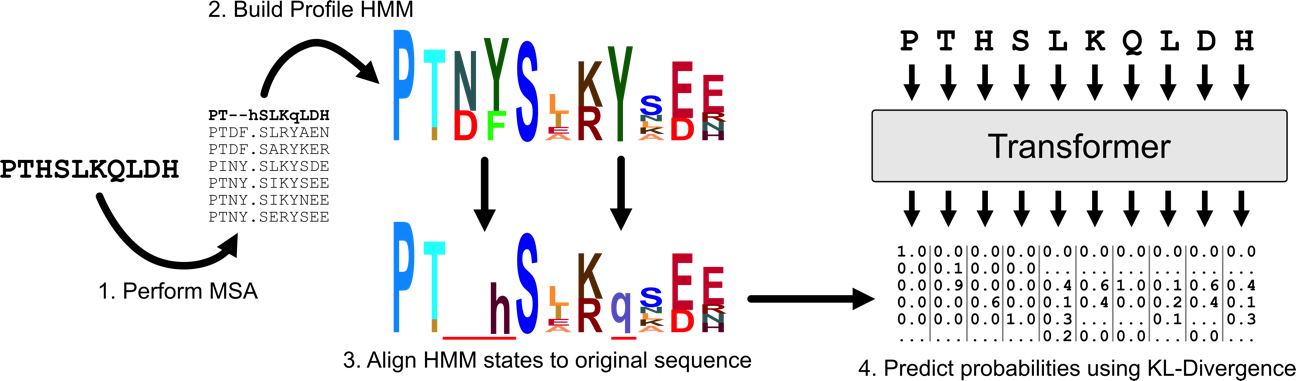}
    \caption{ \small An overview of our proposed task. We start with an initial sequence ``PTHSLKQLDH''. We generate a multiple sequence alignment for that sequence by searching it against a reference database, and we then generate a profile HMM for the alignment. The first H and the Q in our sequence correspond to inserted amino acids that didn't match any columns in the alignment. Therefore, for those amino acids we use insertion state emissions as labels rather than match state emissions. The rest of the amino acids in our sequence were in match states, so we use the match state emission probabilities as labels. Our protein has deletions in two of the match states in the MSA (columns 2 and 3). We omit these from the label since they have no corresponding amino acids as inputs. Finally, we predict the corresponding label using KL divergence, averaged over the length of the sequence.}
    \label{fig:summary_fig}
\end{figure*}

In this paper, we introduce a new pre-training task for protein sequence models. Our task is inspired by alignment-based protein structure predictors: the very models that outperform existing deep NLP models at protein structure prediction \citep{klausen2019netsurfp, elnaggar2020prottrans, ma2015protein}. Such models take as input features derived from multiple sequence alignments (MSAs), which cluster proteins with related sequences. Features derived from MSAs, such as position-specific scoring matrices and hidden Markov model profiles, have long known to be useful features for predicting the structure of a protein \citep{cuff1999evaluation, jones1999protein, cuff2000application, rost2001protein}. Our task posits the reverse: that, in order to predict profiles derived from MSAs from a single protein in the alignment, a network must learn information about that protein's structure. Specifically, we propose to use HMM profiles derived from MSAs as labels during pre-training, rather than as input features in a downstream task. An overview of the task is depicted in Figure \ref{fig:summary_fig}.

The remainder of this paper is organized as follows. 
In Section \ref{sec:mt}, we describe our pre-training task in detail as well as the architecture and hyperparameters we use for further experiments. In Section \ref{sec:rs}, we compare our pre-training task against masked language modeling and a multitask loss that combines the two of them, while keeping architecture, hyperparameters and pre-training data constant. We also compare our model to existing models in the literature. Section \ref{sec:cn} concludes with possibilities for future work.

\section{Methods} \label{sec:mt}
In this section, we describe in detail our pre-training task as well as specifics regarding training and architecture details. We begin with an overview on multiple sequence alignment.

\subsection{Multiple Sequence Alignments}
Multiple sequence alignment (MSA) is a method for grouping a series of related proteins. MSA arranges proteins in a matrix whose rows are individual protein sequences and whose columns contain amino acids that either come from the same position in some ancestral sequence (homologous), or play a common structural or functional role. Building an MSA is a common first step in many downstream tasks including predicting structure and finding homologous proteins \citep{edgar2006multiple}. For our models, we use the pre-training data introduced in \cite{rao2019evaluating}, which comprises 32 million sequences from Pfam \citep{el2019pfam}. Pfam further contains pre-built MSAs for each of its entries, grouped into a set of families. Although it would be possible to build a set of multiple sequence alignments for any protein sequence dataset using standard alignment tools \citep{higgins1992clustal, corpet1988multiple, sievers2011fast}, we use the existing multiple sequence alignments from the 32.0 release of Pfam. 

\subsection{Predicting Alignment Profiles}
Once an MSA is built, it is common to fit a profile HMM, which model the probabilities of amino acids appearing in the columns of an MSA, as well as the probability of inserting additional amino acids between columns or deleting existing columns. It is common to use features derived from profile HMMs as input to structure prediction models, as these profiles contain information about the evolutionary history of a protein \citep{klausen2019netsurfp}. In particular, the emission probabilities give insight into which positions in the proteins are likely to mutate or remain constant over the course of evolution. This in turn illuminates which portions of the protein are critical for the protein's structure or function. For a review on profile HMMs, see \cite{eddy1998profile}. We build profile HMMs from multiple sequence alignments using HMMER \citep{finn2011hmmer} with the default arguments.

Our proposed task is to predict a protein's profile HMM directly from it sequence. Formally, we represent a protein sequence $x$ of length $n$ as $x_1x_2\dots x_n$ and the MSA it belongs to as a matrix:
\[ A = \begin{pmatrix}
a_{11} & a_{12} & \cdots & a_{1m} \\
a_{21} & a_{22} & \cdots & a_{2m} \\
\vdots & \vdots & \cdots & \vdots \\
a_{k1} & a_{k2} & \cdots & a_{km} \\
\end{pmatrix} \]

where $k$ is the number of sequences in the alignment and $m \geq n$ is the length of the alignment. Without loss of generality, we assume that $x$ is the first sequence in the alignment; that is, there exists an injective map $g: [n] \mapsto [m]$ such that $i \leq g(i)$ and $x_i = a_{1g(i)}$ for all $i \in [n]$. We let $h: \{a_{ij} \in A\} \mapsto \{M, I, D\}$ be the MSA state function which maps amino acids to the three possible states in an MSA:
\begin{enumerate}
    \item Match: $a_{ij}$ is an amino acid that is related, evolutionarily or structurally, to other amino acids in column $j$.
    \item Insertion: $a_{ij}$ is an amino acid that is not related to other amino acids in its column, but is more likely the result of a mutation that inserted additional amino acids.
    \item Deletion: $a_{ij}$ is not an amino acid, but rather a column in which protein $i$ is missing an amino acid where other proteins in the MSA have amino acids that are either matched or inserted.
\end{enumerate}
A profile HMM built from this MSA is represented by the match state emissions $p^M_1 p^M_2 \dots p^M_\ell$ and the insertion state emissions $p^I_1 p^I_2 \dots p^I_\ell$, as well as an injective function $f:[\ell] \mapsto [m]$ which maps the indices of the profile back to the columns of the MSA. $p^{M}_j$ and $p^{I}_j$ are probability vectors of size $S$ containing the probability of seeing each amino acid in column $f(j)$ in match or insertions states respectively:
\[\sum_{s=1}^{S} (p^{M}_j)_s = 1, (p^{M}_j)_s \geq 0 \textrm{ and } \sum_{s=1}^{S} (p^{I}_j)_s = 1, (p^{I}_j)_s \geq 0\]

for an amino acid alphabet of size $S$. In our experiments, we use the standard 20 amino acids during profile creation. Note also that $f$ has a well-defined inverse $f^{-1} : [m] \mapsto [\ell]$. 

We define a sequence of vector labels, $l_1 l_2 \dots l_n$ associated with the input sequence $x$, defined as: 
\[l_i(x) = \begin{cases}
        p^M_{f^{-1}(g(i))} & \textrm{ if } h(a_{1g(i)}) = M \\
        p^I_{f^{-1}(g(i))} & \textrm{ if } h(a_{1g(i)}) = I \\
        \end {cases}\]
The $l_i(x)$ are well-defined: $h(a_{1g(i)}) \neq D, \forall i$ since $g(i)$ only maps to columns in the alignment where $x$ contains amino acids. Given a network function $F$, our proposal task's loss function can be written as follows:
\[L_{\textrm{PP}}(x, \theta) = \frac{1}{n}\sum_{i=1}^n \sum_{s=1}^S l_{i, s}(x) (\log (l_{i, s}(x)) - \log (F_{i, s}(x; \theta))) = \frac{1}{n} \sum_{i=1}^n \textrm{KLDiv}(l_i(x), F_i(x; \theta))
\]

where for $F_{i, s}(x; \theta)$ and $l_{i, s}$ the $i$ indexes into the sequence position and the $s$ indexes into the amino acid output probability. The process is described graphically in Figure \ref{fig:summary_fig}. We contrast this loss with the standard masked language modeling objective:

\[L_{\textrm{MLM}}(x, \theta) = \frac{1}{n}\sum_{i \in \textrm{mask}} \sum_{s=1}^S L_{i, s}(x) \log(F_{i, s}(x; \theta)) = \frac{1}{n} \sum_{i \in \textrm{mask}} \textrm{CrossEntropy}(L_i(x), F_i(x; \theta)) \]

for one-hot labels $L_{i, s}(x)$ that are equal to $1$ if $x_i$ is the $s$th amino-acid in the vocabulary, and 0 otherwise. We additionally introduce the joint loss:

\[L_{\textrm{JOINT}}(x, \theta, \lambda) = \lambda L_{\textrm{MLM}}(x, \theta) + (1 - \lambda) L_{\textrm{PP}}(x, \theta) \]

for a scaling parameter $\lambda$, which in practice we set such that $L_{\textrm{MLM}}(x, \theta) \approx L_{\textrm{PP}}(x, \theta)$ over the course of training.

We note here that our task $L_{PP}$ is non-trivial. From an NLP perspective, it is akin to predicting a distribution over possible ways to rephrase a sentence while preserving its meaning from only the original sentence itself. This requires not only knowing which words carry the meaning of the sentence but also knowing the synonyms of these words in the context of that sentence. Doing so would require a significant understanding of language. As such, we believe our task encourages the transformer to learn about underlying protein biology more than simply predicting masked-out amino acids.

\subsection{Data}
To evaluate our pre-training task we make use of the TAPE benchmark: a set of five standardized protein sequence prediction tasks with associated datasets plus a large unlabeled pre-training dataset derived from Pfam \citep{rao2019evaluating}. We build labels for the pre-training data set using the procedure described above. We then evaluate our pre-trained models on the five downstream TAPE tasks: secondary structure prediction \citep{klausen2019netsurfp}, contact prediction \citep{alquraishi2019proteinnet}, remote homology detection \citep{hou2018deepsf}, fluorescence prediction \citep{sarkisyan2016local}, and stability prediction \citep{rocklin2017global}, using the metrics specified by TAPE. 

\subsection{Hyperparameters and Training Details}

For all experiments we train the default transformer architecture used by \cite{rao2019evaluating}, but we note that our pre-training task is not architecture-specific. We train three models with the three different objectives above. The profile prediction model used a learning rate of 0.00025, while the multi-task and masked language modeling models use a learning rate of 0.0001. These learning rates represented the largest learning rates that did not cause the model to diverge during the course of training, searching from 0.00001 in increments of 0.00005. All models were pre-trained for 34 epochs.\footnote{We cut off pre-training at 34 epochs due to time constraints. It is possible that further gains may result from continuing to pre-train for longer.} The learning rate uses a warm-up schedule and dynamic batch sizing, both of which are described in \cite{rao2019evaluating}. Pre-training a single model took approximately two weeks with 8 NVIDIA Tesla V100 GPUs.

Training details for all downstream tasks follow the procedure laid out by \cite{rao2019evaluating}: namely, a learning rate of 0.0001 with linear warm-up schedule, the Adam optimizer and backpropagation through the entire pre-trained model. The downstream prediction heads all follow those in \cite{rao2019evaluating}, except for contact prediction which uses a single linear layer rather than a 30-layer convolutional architecture.\footnote{This new linear prediction head is the default in the current version of the \texttt{tape-proteins} package.} 

\section{Results} \label{sec:rs}

\subsection{Comparing Pre-training Tasks}
We first compare our pre-training task against masked language modeling and the multitask model which combines both tasks, keeping hyperparameters and architecture fixed. The results are shown in Tables~\ref{tab:comp-structure},~\ref{tab:comp-homology},  and \ref{tab:comp-engineering}. For both structure prediction tasks---secondary structure and contact prediction---profile prediction pre-training outperforms multitasking, which in turn outperforms masked language modeling. All three tasks outperform the same model that was not pre-trained. Although it is not surprising that profile pre-training outperforms mask language modeling on structure prediction---namely because HMM profiles are known to contain information relevant to a protein's structure---the differences between the evaluated models are not large. This may mean that potentially more than just a new pre-training task is needed to continue to improve structure predictors, such as different architectures, or larger pre-training datasets \citep{elnaggar2020prottrans}.

\begin{table}
\caption{Comparing Downstream Performance on Structure Prediction}
\small
    \centering
    \begin{tabular}{lcccc}
        \toprule
        \multicolumn{1}{c}{Pretraining Task} &\multicolumn{3}{c}{Secondary Structure} &
        \multicolumn{1}{c}{Contact Prediction}\\
        \cmidrule(l){2-4} \cmidrule(l){5-5}
                    & CASP12 & CB513 & TS115 & CASP12 \\
        \midrule
        Profile Prediction       & \textbf{0.71} & \textbf{0.74} & \textbf{0.77} & \textbf{0.33}\\
        Masked Language Modeling & 0.67 & 0.72 & 0.75 & 0.24\\
        Multi-Task               & \textbf{0.71} & 0.72 & 0.76 & 0.28  \\
        No Pretraining           & 0.67 & 0.64 & 0.68 & 0.05\\
        \bottomrule
    \end{tabular}
    \label{tab:comp-structure}
\end{table}

The remote homology detection task demonstrates the largest gap between profile prediction and mask language modeling. The model pre-trained with profile prediction is about 2 to 3 times more accurate than the model pre-trained using masked language modeling. The performance of the multitask model lies between that of the other two models and all three again outperform a randomly initialized model. This may be because HMM profiles also contain significant amounts of information about evolutionarily related proteins, which is closely related to the structural or functional groupings that a protein falls into.

\begin{table}
\caption{Comparing Downstream Performance on Homology Detection}
\small
    \centering
    \begin{tabular}{lccc}
        \toprule
        \multicolumn{1}{c}{Pretraining Task}
                    & Fold & Superfamily & Family \\
        \midrule
        Profile Prediction       & \textbf{0.23} & \textbf{0.45} & \textbf{0.86}\\
        Masked Language Modeling & 0.14 & 0.16 & 0.45\\
        Multi-Task               & 0.16 & 0.28 & 0.68\\
        No Pretraining           & 0.05 & 0.05 & 0.21\\
        \bottomrule
    \end{tabular}
    \label{tab:comp-homology}
\end{table}

We find the same pattern on the fluorescence task: profile prediction leads to the best test set performance, followed by multitasking, masked language modeling and no pre-training in that order. Finally, on the stability task, the masked language modeling model and the multitask model both outperform profile prediction. This may be because this task tests models' ability to generalize to proteins with a single amino acid difference from proteins in the training set---a task that masked language modeling is particularly suited for. Taken as a whole, these results indicate that there may not be a one-size-fits-all pre-training task for all downstream prediction tasks. Rather, it may be beneficial to tailor the pre-training task to the downstream task: for structure or evolutionary tasks, incorporating profile information may be beneficial, but for fine-grained engineering tasks, masked language modeling may be a better choice.

\begin{table}
\caption{Comparing Downstream Performance on Engineering Tasks}
\small
    \centering
    \begin{tabular}{lcc}
        \toprule
        \multicolumn{1}{c}{Pretraining Task} & \multicolumn{2}{c}{Downstream Task}\\
        \cmidrule(l){2-3}
                    & Fluorescence & Stability \\
        \midrule
        Profile Prediction       & \textbf{0.38} & 0.55 \\
        Masked Language Modeling & 0.25 & \textbf{0.63} \\
        Multi-Task               & 0.28 & \textbf{0.63} \\
        No Pretraining           & 0.04 & 0.45 \\
        \bottomrule
    \end{tabular}
    \label{tab:comp-engineering}
\end{table}

\subsection{Comparing Against TAPE Benchmarks}
In this section, we compare against the models presented in the original TAPE benchmark, as well as some previous literature that makes use of the TAPE benchmark. We omit the contact prediction task because the \texttt{tape-proteins} package significantly changed the downstream contact prediction architecture after the paper's original publication. We note that this was also done by previous work \citep{Lu2020.09.04.283929}. For secondary structure, we present results from the CB513 test set. For remote homology detection we present results from the fold level prediction task. The results are presented in Table \ref{tab:downstream-results}. 

On the secondary structure task, we fall short of the NetsurfP2.0 model presented by \cite{klausen2019netsurfp}, which is the alignment baseline from \cite{rao2019evaluating}. We also fall short of the LSTM and ResNet models from \cite{rao2019evaluating}, but outperform both the Transformer model as well as all previous work proposing protein-specific pre-training tasks \citep{bepler2018learning, Lu2020.09.04.283929}, and a recent paper employing an auto-regressive LSTM \citep{alley2019unified}. On the remote homology task we outperformed all existing models except the TAPE benchmark's LSTM model and the LSTM presented by \citep{alley2019unified}. We again note that we outperform the  protein-specific pre-training tasks in \cite{bepler2018learning} and \cite{Lu2020.09.04.283929}. In the engineering tasks, all of our models are outperformed by existing work, including models from the original TAPE benchmark and the related works mentioned above.

We also note that we do not view most existing work as mutually exclusive with ours. Instead, we believe that combining our task with the architectures and pre-training tasks present in existing work may lead to further performance gains.

\begin{table}
\caption{Results on downstream supervised tasks }
\begin{adjustwidth}{-1in}{-1in}
\small
    \centering
    \begin{tabular}{llcccc}
        \toprule
        \multicolumn{2}{c}{Method} & \multicolumn{1}{c}{Structure} & \multicolumn{1}{c}{Evolutionary} & \multicolumn{2}{c}{Engineering}\\
        \cmidrule(l){3-3} \cmidrule(l){4-4} \cmidrule(l){5-6}
          & & SS & Homology & Fluorescence & Stability \\
        \midrule
        Pre-trained  & Transformer & 0.73& 0.21 &\textbf{0.68} & \textbf{0.73} \\
        models from                           & LSTM        & 0.75 & \textbf{0.26} & 0.67 & 0.69 \\
        \cite{rao2019evaluating}     & ResNet      & 0.75 & 0.17 & 0.21 & \textbf{0.73} \\
                                   \cmidrule(l){2-6}
        Baselines from             & One-hot      & 0.69 & 0.09 & 0.14 & 0.19 \\
        \cite{rao2019evaluating}                   & Alignment & \textbf{0.80} & 0.09 & N/A & N/A \\
                                  \cmidrule(l){2-6}
         & \cite{bepler2018learning}         & 0.73  & 0.17 & 0.33 & 0.64 \\
         & \cite{alley2019unified}           & 0.73  & 0.23 & 0.67 & \textbf{0.73} \\
         & \cite{Lu2020.09.04.283929}        & 0.70   & 0.13 & \textbf{0.68} & 0.68 \\
                                    \cmidrule(l){2-6}
         Our models & Profile Prediction         & 0.74  & 0.23 & 0.38 & 0.55 \\
            &   Masked Language Modeling         & 0.72 & 0.14 & 0.25 & 0.63 \\
         & Multi-Task                 & 0.74 & 0.16 & 0.28 & 0.63 \\
        \bottomrule
    \end{tabular}
         \label{tab:downstream-results}

\end{adjustwidth}

\end{table}

\section{Conclusion} \label{sec:cn}
In this work, we produced a new pre-training task---profile prediction---which encourages a transformer to learn a distribution over evolutionarily-related protein sequences from a single input  sequence. Our task alone outperforms masked language modeling on  downstream tasks related to structure and homology detection. However, we do find that mask language modeling is still a useful pre-training task for engineering-related downstream tasks, namely, predicting intrinsic stability. We believe that our work shows that pre-training tasks borrowed from natural language processing models may not always be ideal for protein sequence models. Instead, we hope that these results inspire additional interest in developing biologically-informed protein sequence pre-training tasks. The combination of novel pre-training tasks, new architectures, and massively unlabeled datasets may be the key to eventually outperforming alignment-based predictors.

\bibliography{neurips_2020}
\bibliographystyle{acl_natbib}

\end{document}